\pdfoutput=1
\documentclass[11pt,a4paper]{article}
\usepackage[hyperref]{acl2020}
\usepackage{times}
\usepackage{latexsym}

\usepackage{microtype}

\aclfinalcopy %

\usepackage[T1]{fontenc}
\usepackage{multirow}
\usepackage{multicol}
\usepackage{graphicx}
\usepackage{subcaption}
\usepackage{mathtools}
\usepackage{listings}
\usepackage{tikz}
\usetikzlibrary{positioning}
\usepackage{soul}
\usepackage{xspace}

\definecolor{gray}{RGB}{211,211,211}
\newcommand{\jbasicstyle}{\small\sffamily} %

\newcommand{\jnumberstyle}{\scriptsize}

\makeatletter
\newenvironment{btHighlight}[1][]
{\begingroup\tikzset{bt@Highlight@par/.style={#1}}\begin{lrbox}{\@tempboxa}}
{\end{lrbox}\bt@HL@box[bt@Highlight@par]{\@tempboxa}\endgroup}

\newcommand\btHL[1][]{%
  \begin{btHighlight}[#1]\bgroup\aftergroup\bt@HL@endenv%
}
\def\bt@HL@endenv{%
  \end{btHighlight}%
  \egroup
}
\newcommand{\bt@HL@box}[2][]{%
  \tikz[#1]{%
    \pgfpathrectangle{\pgfpoint{1pt}{0pt}}{\pgfpoint{\wd #2}{\ht #2}}%
    \pgfusepath{use as bounding box}%
    \node[anchor=base west, fill=orange!30,outer sep=0pt,inner xsep=1pt, inner ysep=0pt, rounded corners=1pt, minimum height=\ht\strutbox,#1]{\raisebox{1pt}{\strut}\strut\usebox{#2}};
  }%
}

\newenvironment{btHighlightLine}[1][]
{\begingroup\tikzset{bt@HighlightLine@par/.style={#1}}\begin{lrbox}{\@tempboxa}}
{\end{lrbox}\bt@HLLine@box[bt@HighlightLine@par]{\@tempboxa}\endgroup}

\newcommand\btHLLine[1][]{%
  \begin{btHighlightLine}[#1]\bgroup\aftergroup\bt@HLLine@endenv%
}
\def\bt@HLLine@endenv{%
  \end{btHighlightLine}%
  \egroup
}
\newcommand{\bt@HLLine@box}[2][]{%
  \tikz[#1]{%
    \pgfpathrectangle{\pgfpoint{0pt}{-1pt}}{\pgfpoint{\wd #2}{\ht #2}}%
    \pgfusepath{use as bounding box}%
    \node[anchor=base west, fill=orange!30,outer sep=0pt,inner xsep=0pt, inner ysep=0pt, minimum height=\ht\strutbox+3pt, minimum width=\linewidth,#1] (line-bg) {};
    \node[right = 0 of line-bg.west, outer sep=0pt, inner xsep=0pt, inner ysep=0pt]{\raisebox{0pt}{\strut}\strut\usebox{#2}};
  }%
}
\makeatother

\makeatletter
\newcommand\BeraMonottfamily{%
  \def\fvm@Scale{0.85}%
    \fontfamily{fvm}\selectfont%
  }
\makeatother

\lstdefinelanguage{pseudo}
{
  morekeywords={},
  keywordstyle=\bfseries,
  lineskip=-0.1em,
  numbers=left, %
  numberstyle=\jnumberstyle,
  numbersep=4pt,
  basicstyle=\jbasicstyle,
  breaklines=true,
  breakautoindent=true,
  tabsize=2,
  columns=fullflexible,
  morecomment=*[l][\textsl]{//},
  mathescape=true,
  xleftmargin=10pt,
}

\lstdefinelanguage{todo-comment}
{
  morekeywords={},
  keywordstyle=\bfseries,
  lineskip=-0.1em,
  numbers=none,
  basicstyle=\jbasicstyle,
  breaklines=true,
  breakautoindent=true,
  tabsize=2,
  columns=fullflexible,
  morecomment=*[l][\textsl]{//},
  mathescape=true,
  xleftmargin=-10pt,
}

\lstdefinelanguage{java-pretty}
{
  language=java,
  numbers=none,
  basicstyle=\scriptsize\BeraMonottfamily,
  numberstyle=\scriptsize,
  breaklines=true,
  showstringspaces=false,
  moredelim=**[is][\color{red!60!black}]{(-W<)}{(>W-)},
  moredelim=**[is][\color{green!60!black}]{(+W<)}{(>W+)},
  literate=
         {-}{-}{1}
}

\lstdefinelanguage{java-diff}
{
  language=java,
  numbers=none,
  basicstyle=\tiny\BeraMonottfamily,
  numberstyle=\tiny,
  breaklines=true,
  showstringspaces=false,
  moredelim=**[il][{\btHLLine[fill=red!10]}]{(-L)},
  moredelim=**[il][{\btHLLine[fill=green!10]}]{(+L)},
  moredelim=**[is][{\btHL[fill=red!20]}]{(-W<)}{(>W-)},
  moredelim=**[is][{\btHL[fill=green!20]}]{(+W<)}{(>W+)},
  literate=
         {-}{-}{1}
}

\lstdefinelanguage{return-comment}
{
  language=,
  columns=fixed,
  keepspaces=true,
  numbers=none,
  basicstyle=\tiny\BeraMonottfamily,
  numberstyle=\tiny,
  breaklines=true,
  showstringspaces=false,
  moredelim=**[il][{\btHLLine[fill=red!10]}]{(-L)},
  moredelim=**[il][{\btHLLine[fill=green!10]}]{(+L)},
  moredelim=**[is][{\btHL[fill=red!20]}]{(-W<)}{(>W-)},
  moredelim=**[is][{\btHL[fill=green!20]}]{(+W<)}{(>W+)},
}

\newcommand{\nocell}[1]{\multicolumn{#1}{c|}{}}

\newcommand{\DefMacro}[2]{\expandafter\newcommand\csname rmk-#1\endcsname{#2}}
\newcommand{\UseMacro}[1]{\csname rmk-#1\endcsname}

\newcommand{\Return}{\texttt{@return}}
\newcommand{\OldComment}{C\textsubscript{old}}
\newcommand{\NewComment}{C\textsubscript{new}}
\newcommand{\PredictedNewComment}{C'\textsubscript{new}}
\newcommand{\EditComment}{C\textsubscript{edit}}
\newcommand{\PredictedEditComment}{C'\textsubscript{edit}}
\newcommand{\OldCode}{M\textsubscript{old}}
\newcommand{\NewCode}{M\textsubscript{new}}
\newcommand{\EditCode}{M\textsubscript{edit}}
\newcommand{\PointerEditComment}{P\textsubscript{edit}}
\newcommand{\PointerOldComment}{P\textsubscript{old}}

\newcommand{\Insert}{\texttt{\small Insert}}
\newcommand{\Delete}{\texttt{\small Delete}}
\newcommand{\Keep}{\texttt{\small Keep}}
\newcommand{\Replace}{\texttt{\small Replace}}
\newcommand{\ReplaceOld}{\texttt{\small ReplaceOld}}
\newcommand{\ReplaceNew}{\texttt{\small ReplaceNew}}
\newcommand{\ReplaceKeepBefore}{\texttt{\small ReplaceKeepBefore}}
\newcommand{\ReplaceKeepAfter}{\texttt{\small ReplaceKeepAfter}}
\newcommand{\DeleteOld}{\texttt{\small DeleteOld}}
\newcommand{\DeleteNew}{\texttt{\small DeleteNew}}
\newcommand{\DeleteKeepBefore}{\texttt{\small DeleteKeepBefore}}
\newcommand{\DeleteKeepAfter}{\texttt{\small DeleteKeepAfter}}
\newcommand{\InsertOld}{\texttt{\small InsertOld}}
\newcommand{\InsertNew}{\texttt{\small InsertNew}}
\newcommand{\InsertKeepBefore}{\texttt{\small InsertKeepBefore}}
\newcommand{\InsertKeepAfter}{\texttt{\small InsertKeepAfter}}
\newcommand{\ReplaceOldKeepBefore}{\texttt{\small ReplaceOldKeepBefore}}
\newcommand{\ReplaceNewKeepBefore}{\texttt{\small ReplaceNewKeepBefore}}
\newcommand{\ReplaceOldKeepAfter}{\texttt{\small ReplaceOldKeepAfter}}
\newcommand{\ReplaceNewKeepAfter}{\texttt{\small ReplaceNewKeepAfter}}
\newcommand{\InsertOldKeepBefore}{\texttt{\small InsertOldKeepBefore}}
\newcommand{\InsertNewKeepBefore}{\texttt{\small InsertNewKeepBefore}}
\newcommand{\InsertOldKeepAfter}{\texttt{\small InsertOldKeepAfter}}
\newcommand{\InsertNewKeepAfter}{\texttt{\small InsertNewKeepAfter}}
\newcommand{\DeleteOldKeepBefore}{\texttt{\small DeleteOldKeepBefore}}
\newcommand{\DeleteNewKeepBefore}{\texttt{\small DeleteNewKeepBefore}}
\newcommand{\DeleteOldKeepAfter}{\texttt{\small DeleteOldKeepAfter}}
\newcommand{\DeleteNewKeepAfter}{\texttt{\small DeleteNewKeepAfter}}

\newcommand{\editmodel}{edit model\xspace}
\newcommand{\EditModel}{Edit Model\xspace}
\newcommand{\editmodels}{edit models\xspace}
\newcommand{\editmodeling}{edit modeling\xspace}

\definecolor{MyGreen}{RGB}{34,139,34}

\def\@fnsymbol#1{\ensuremath{\ifcase#1\or *\or \dagger\or \ddagger\or
   \mathsection\or \mathparagraph\or \|\or **\or \dagger\dagger
   \or \ddagger\ddagger \else\@ctrerr\fi}}
\newcommand{\ssymbol}[1]{^{\@fnsymbol{#1}}}

\newcommand{\CorpusSize}{7,239}
\newcommand{\NumProjects}{1,081}
\newcommand{\CodeVocabSize}{5,945}
\newcommand{\CommentVocabSize}{3,642}
\newcommand{\GenTrainSet}{103,473}
\newcommand{\TestSet}{736}

\newcommand{\BeamWidth}{20}

\title{Learning to Update Natural Language Comments Based on Code Changes}

\author{
Sheena Panthaplackel\textsuperscript{\rm 1},
Pengyu Nie\textsuperscript{\rm 2},
Milos Gligoric\textsuperscript{\rm 2},
Junyi Jessy Li\textsuperscript{\rm 3},
Raymond J. Mooney\textsuperscript{\rm 1}\\
\textsuperscript{\rm 1}Department of Computer Science\\
\textsuperscript{\rm 2}Department of Electrical and Computer Engineering\\
\textsuperscript{\rm 3}Department of Linguistics\\
The University of Texas at Austin\\
\texttt{spantha@cs.utexas.edu, pynie@utexas.edu, gligoric@utexas.edu,}\\ \texttt{jessy@austin.utexas.edu, mooney@cs.utexas.edu}
}

\date{}

\begin{document}
\maketitle

\begin{abstract}
We formulate the novel task of automatically updating an existing natural language comment based on changes in the body of code it accompanies. We propose an approach that learns to correlate changes across two distinct language representations, to generate a sequence of edits that are applied to the existing comment to reflect the source code modifications. We train and evaluate our model using a dataset that we collected from commit histories of open-source software projects, with each example consisting of a concurrent update to a method and its corresponding comment. We compare our approach against multiple baselines using both automatic metrics and human evaluation. Results reflect the challenge of this task and that our model outperforms baselines with respect to making edits.
\end{abstract}

\section{Introduction}

Software developers include natural language comments alongside source code as a way to document various aspects of the code such as functionality, use cases, pre-conditions, and post-conditions. With the growing popularity of open-source software that is widely used and jointly developed, the need for efficient communication among developers about code details has increased. Consequently, comments have assumed a vital role in the development cycle. 
With developers regularly refactoring and iteratively incorporating new functionality, source code is constantly evolving; however, the accompanying comments are not always updated to reflect the code changes~\cite{icomment2007,ratol2017fragile}. Inconsistency between code and comments can not only lead time-wasting confusion in tight project schedules~\cite{Hu2018DeepCC} but can also result in bugs~\cite{icomment2007}. 
To address this problem, we propose an approach that can automatically suggest comment updates when the associated methods are changed.

Prior work explored rule-based approaches for detecting inconsistencies for a limited set of cases; however, they do not present ways to automatically fix these inconsistencies~\cite{icomment2007,ratol2017fragile}. Recent work in automatic comment generation aims to generate a comment given a code representation~\cite{Liang2018AutomaticGenerationComments, Hu2018DeepCC, Fernandes2019StructuredNeural}; although these techniques could be used to produce a completely new comment that corresponds to the most recent version of the code, this could potentially discard salient content from the existing comment that should be retained. To the best of our knowledge, we are the first to formulate the task of \emph{automatically updating an existing comment when the corresponding body of code is modified}.

\newsavebox\boxOldGetRotX
\begin{lrbox}{\boxOldGetRotX}
  \begin{lstlisting}[language=java-pretty]
(*@{\sffamily /**{\ttfamily\bf @return} double the roll euler angle.*/}@*)
public double getRotX() {
    return mOrientation.getRotationX();
}
  \end{lstlisting}
\end{lrbox}

\newsavebox\boxNewGetRotX
\begin{lrbox}{\boxNewGetRotX}
  \begin{lstlisting}[language=java-pretty]
(*@{\sffamily /**{\ttfamily\bf @return} double the roll euler angle {\color{green!60!black}in degrees}.*/}@*)
public double getRotX() {
    return (+W<)Math.toDegrees((>W+)mOrientation.getRotationX()(+W<))(>W+);
}
  \end{lstlisting}
\end{lrbox}

\begin{figure}[t]
\centering
\begin{tikzpicture}[
  line width=0.4pt,
  node distance=0ex and 0em,
  every node/.style={scale=1},
  gridBox/.style={rectangle, opacity=0, draw=red},
  box/.style={rectangle, draw=black, inner sep=2pt, font=\small},
  rounded box/.style={rectangle, rounded corners, draw=black, inner sep=2pt, font=\small},
  anno/.style={font=\footnotesize},
]

  \DefMacro{wCodeBox}{20em}
  \DefMacro{hCodeBox}{7.5ex}
  
  \node (g-Old) at (0,0) [box, minimum width=\UseMacro{wCodeBox}, minimum height=\UseMacro{hCodeBox}] {};
  \node (b-Old) [right = 0 of g-Old.west] [] {\usebox\boxOldGetRotX};
  \node (b-OldText) [below left = 1pt and 1pt of g-Old.north east] [box, draw=red!50, fill=red!20, scale=0.8] {\bf Previous Version\phantom{p\hskip -.6em}};
  \node (g-New) [below = 0 of g-Old.south] [box, minimum width=\UseMacro{wCodeBox}, minimum height=\UseMacro{hCodeBox}] {};
  \node (b-New) [right = 0 of g-New.west] [] {\usebox\boxNewGetRotX};
  \node (b-NewText) [below left = 1pt and 1pt of g-New.north east] [box, draw=green!50, fill=green!20, scale=0.8] {\bf Updated Version};
  
\end{tikzpicture}
\vspace{-18pt}
\caption{\small Changes in the \texttt{getRotX} method  and its corresponding  \Return{} comment between two subsequent commits of the rajawali-rajawali project, available on GitHub.}
\label{fig:rajawali}
\vspace{-15pt}
\end{figure}

This task is intended to align with how
developers edit a comment when they introduce changes in the corresponding method.
Rather than deleting it and starting from scratch, they would likely only modify the specific parts relevant to the code updates. For example, Figure~\ref{fig:rajawali} shows the \texttt{getRotX} method being modified to have the return value parsed into degrees. Within the same commit, the corresponding comment is revised to indicate this, without imposing changes on parts of the comment that pertain to other aspects of the return value. 
We replicate this process through a novel approach which is designed to correlate edits across two distinct language representations: source code and natural language comments. Namely, our model is trained to generate a sequence of \emph{edit actions}, which are to be applied to the existing comment, by conditioning on learned representations of the code edits and existing comment. 
We additionally incorporate linguistic and lexical features to guide the model in determining where edits should be made in the existing comment. Furthermore, we develop an output reranking scheme that aims to produce edited comments that are fluent, preserve content that should not be changed, and maintain stylistic properties of the existing comment.

We train and evaluate our system on a corpus constructed from open-source Java projects on GitHub, by mining their commit histories and extracting examples from consecutive commits in which there was a change to both the code within a method as well as the corresponding Javadoc comment, specifically, the \Return{} Javadoc tag. These comments, which have been previously studied for learning associations between comment and code entities~\cite{panthaplackel2020associating}, follow a well-defined structure and describe characteristics of the output of a method. For this reason, as an initial step, we focus on \Return{} comments in this work.
Our evaluation consists of several automatic metrics that are used to evaluate language generation tasks as well as tasks that relate to editing natural language text. We also conduct human evaluation, and assess whether human judgments correlate with the automatic metrics.

The main contributions of this work include (1)~the task of automatically updating an existing comment based on source code changes and (2)~a novel approach for learning to relate edits between source code and natural language that outperforms multiple baselines on several automatic metrics and human evaluation. Our implementation and data are publicly available.\footnote{\url{https://github.com/panthap2/LearningToUpdateNLComments}}

\section{Task}
Given a method, its corresponding comment, and an updated version of the method, the task is to update the comment so that it is consistent with the code in the new method. For the example in Figure~\ref{fig:rajawali}, we want to generate \emph{``\Return{} double the roll euler angle in degrees.''} based on the changes between the two versions of the method and the existing comment \emph{``\Return{} double the roll euler angle.''} Concretely, given (\OldCode{}, \OldComment{}) and \NewCode{}, where \OldCode{} and \NewCode{} denote the old and new versions of the method, and \OldComment{} signifies the previous version of the comment, the task is to produce \NewComment{}, the updated version of the comment.

\section{\EditModel Overview}
\label{sec:edit-model}
We design a system that examines source code changes and how they relate to the existing comment in order to produce an updated comment that reflects the code modifications. Since \OldComment{} and \NewComment{} are closely related, training a model to directly generate \NewComment{} risks having it learn to just copy \OldComment{}. 
To explicitly inform the model of edits,
we define the target output as a \emph{sequence of edit actions}, \EditComment{}, to indicate how the existing comment should be revised (e.g., for \OldComment{}=\texttt{ABC},
\EditComment{}=$\texttt{\small<Delete>A<DeleteEnd>}$
implies that \texttt{A} should be deleted to produce \NewComment{}=\texttt{BC}). Furthermore, in order to better correlate these edits with changes in the code, we unify \OldCode{} and \NewCode{} into a single \textit{diff} sequence that explicitly identifies code edits, \EditCode{}. We discuss in more detail how \EditCode{} and the training \EditComment{} are constructed in \S\ref{sec:edits}.

\begin{figure}[t]
\centering
\includegraphics[width=\columnwidth]{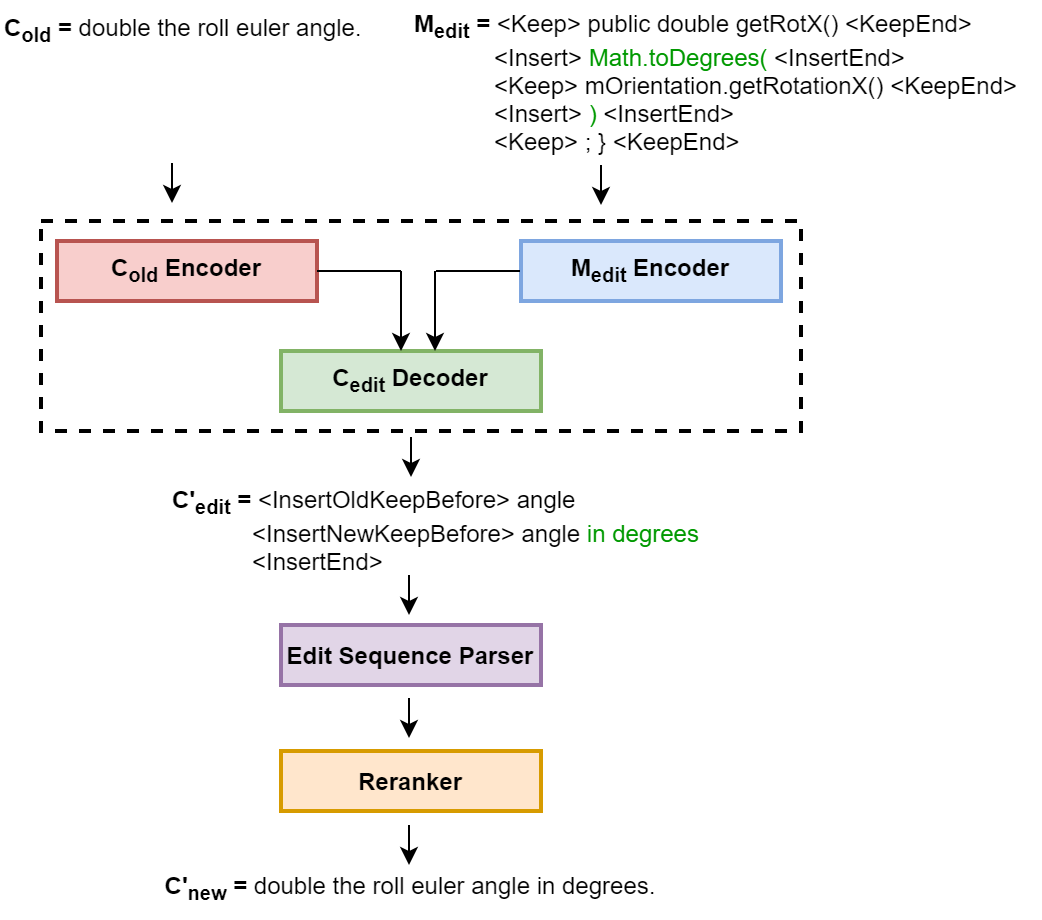}
\vspace{-5pt}
\caption{\small High-level overview of our system.}
\label{fig:architecture}
\vspace{-10pt}
\end{figure}

Figure~\ref{fig:architecture} shows a high-level overview of our system. We design an 
encoder-decoder architecture
consisting of three components: a two-layer, bi-directional GRU~\cite{ChoGRU} that encodes the code changes (\EditCode{}), another two-layer, bi-directional GRU that encodes the existing comment (\OldComment{}), and a GRU that is trained to decode a sequence of edit actions (\EditComment{}).\footnote{We refrain from using the self-attention model~\cite{transformer} because prior work~\cite{Fernandes2019StructuredNeural} suggests that it yields lower performance for comment generation.} 
We concatenate the final states of the two encoders to form a vector that summarizes the content in \EditCode{} and \OldComment{}, and use this vector as the initial state of the decoder. The decoder essentially has three subtasks: (1) identify edit locations in \OldComment{}; (2) determine parts of \EditCode{} that pertain to making these edits; and (3) apply updates in the given locations based on the relevant code changes. We rely on an attention mechanism~\cite{Luong2015Attention} over the hidden states of the two encoders to accomplish the first two goals. At every decoding step, rather than aligning the current decoder state with all the encoder hidden states jointly, we align it with the hidden states of the two encoders separately. We concatenate the two resulting context vectors to form a unified context vector that is used in the final step of computing attention, ensuring that we incorporate pertinent content from both input sequences. Consequently, the resulting attention vector carries information relating to the current decoder state as well as knowledge aggregated from relevant portions of \OldComment{} and \EditCode{}. 

Using this information, the decoder performs the third subtask, which requires reasoning across language representations. Specifically, it must determine how the source code changes that are relevant to the current decoding step should manifest as natural language updates to the relevant portions of \OldComment{}. At each step, it decides whether it should begin a new edit action by generating an edit start keyword, continue the present action by generating a comment token, or terminate the present action by generating an end-edit keyword. Because actions relating to deletions will include tokens in \OldComment{}, and actions relating to insertions are likely to include tokens in \EditCode{}, we equip the decoder with a pointer network~\cite{VinyalsPointer} to accommodate copying tokens from \OldComment{} and \EditCode{}. The decoder generates a sequence of edit actions, which will have to be parsed into a comment (\S\ref{sec:post-processing}).

\section{Representing Edits}
\label{sec:edits}
Here we define the edit lexicon that is used to construct the input code edit sequence, \EditCode{}, and the target comment edit sequence, \EditComment{}.

\subsection{Edit Lexicon}
\label{sec:edit-lexicon}
We use difflib\footnote{\url{https://docs.python.org/3/library/difflib.html}} to extract code edits and target comment edits. Both the input code edit sequence and the target comment edit sequence consist of a series of edit actions; each edit action is structured as 
\texttt{\small<Action> [span of tokens] <ActionEnd>}.\footnote{Preliminary experiments showed that this performed better than structuring edits at the token-level as in other tasks~\cite{ProgramRepairShin, li-style-transfer, dong-sentence-simplification, awasthi-etal-2019-parallel}.}

We define four types of edit actions: \Insert{}, \Delete{}, \Replace{}, and \Keep{}. Because the \Replace{} action must simultaneously incorporate distinct content from two versions (i.e., tokens in the old version that will be replaced, and tokens in the new version that will take their place), it follows a slightly different structure:

\noindent
\begin{equation*}
  \begin{array}{l}
    \texttt{\small <ReplaceOld> [span of old tokens]}\\
    \noalign{\vskip-6pt}
    \texttt{\small <ReplaceNew> [span of new tokens]}\\
    \noalign{\vskip-6pt}
    \texttt{\small <ReplaceEnd>}\\
    \noalign{\vskip-6pt}
  \end{array}
\end{equation*}

\subsection{Code Edits}
We extract the edits between \OldCode{} and \NewCode{} using the edit lexicon to construct \EditCode{}, the code edit sequence used as input in one of the encoders. 
Figure~\ref{fig:architecture} (top right) shows the \EditCode{} corresponding to code changes in Figure~\ref{fig:rajawali}.

In contrast to line-level code \textit{diffs} that are commonly used for commit message generation~\cite{LoyolaETAL17GeneratingDescriptionFromSourceCodeChanges, JiangCommit, XuCommit}, this representation allows us to explicitly capture more fine-grained edits. While we could exploit the abstract syntax tree (AST) structure of source code and represent the changes between the ASTs corresponding to the two versions of code, prior work suggests that such techniques do not always lead to improved performance~\cite{yin19iclr}. We leave it to future work to investigate how the AST structure can be leveraged for this task.

\subsection{Comment Edits}
We identify the changes between \OldComment{} and \NewComment{} to construct \EditComment{}, the target comment edit sequence. During inference, the output comment is produced by parsing the predicted edit sequence (\S\ref{sec:post-processing}).
We introduce a slightly modified set of specifications that disregards the \Keep{} type when constructing the sequence of edit actions, referred to as the \textit{condensed edit sequence}.

The intuition for disregarding \Keep{} and the span of tokens to which it applies is that we can simply copy the content that is retained between \OldComment{} and \NewComment{}, instead of generating it anew.
By doing post-hoc copying, we simplify learning for the model since it has to only learn \textit{what to change} rather than also having to learn \textit{what to keep}.

We design a method to deterministically place edits in their correct positions in the absence of \Keep{} spans. 
For the example in Figure~\ref{fig:rajawali}, the raw sequence 
$\texttt{\small<Insert>in degrees<InsertEnd>}$ does not encode information as to where ``in degrees'' should be inserted. To address this, we bind an insert sequence with the minimum number of words (aka ``anchors'') such that the place of insertion can be uniquely identified. This results in 
the structure that is shown for \EditComment{} in Figure~\ref{fig:architecture}.
Here ``angle'' serves as the anchor point, identifying the insert location. Following the structure of \Replace{}, this sequence indicates that ``angle'' should be replaced with ``angle in degrees,'' effectively inserting ``in degrees'' and keeping ``angle'' from \OldComment{}, which appears immediately before the insert location. 
See Appendix~\ref{appendix:comment-edit-lexicon} for details on this procedure.

\subsection{Parsing Edit Sequences}
\label{sec:post-processing}
Since the decoder is trained to predict a sequence of edit actions, we must align it with \OldComment{} and copy unchanged tokens in order to produce the edited comment. We denote the predicted edit sequence as \PredictedEditComment{} and the corresponding parsed output as \PredictedNewComment{}. 
This procedure entails simultaneously following pointers, left-to-right, on \OldComment{} and \PredictedEditComment{}, which we refer to as \PointerOldComment{} and \PointerEditComment{} respectively. \PointerOldComment{} is advanced, copying the current token into \PredictedNewComment{} at each point, until an edit location is reached. The edit action corresponding to the current position of \PointerEditComment{} is then applied, and the tokens from its relevant span are copied into \PredictedNewComment{} if applicable. Finally, \PointerEditComment{} is advanced to the next action, and \PointerOldComment{} is also advanced to the appropriate position in cases involving deletions and replacements. This process repeats until both pointers reach the end of their respective sequences.

\section{Features}
\label{sec:features}
\vspace{-5pt}
We extract linguistic and lexical features for tokens in \EditCode{} and \EditComment{}, many of which were shown to improve learning associations between \Return{} comment and source code entities in our prior work~\cite{panthaplackel2020associating}. We incorporate these features into the network as one-hot vectors that are concatenated to \EditCode{} and \EditComment{} embeddings
and then passed through a linear layer. These vectors are provided as inputs to the two encoders. 
All sequences are subtokenized, e.g., \texttt{camelCase} $\rightarrow$ \texttt{camel}, \texttt{case}.

\noindent\textbf{Features specific to \EditCode{}:}
We aim to take advantage of common patterns among different types of code tokens by incorporating features that identify certain categories: edit keywords, Java keywords, and operators. If a token is not an edit keyword, we have indicator features for whether it is part of a \Insert{}, \Delete{}, \ReplaceNew{}, \ReplaceOld{}, or \Keep{} span. We believe this will be particularly helpful for longer spans since edit keywords only appear at either the beginning or end of a span.
Finally, we include a feature to indicate whether the token matches a token in \OldComment{}. This is intended to help the model identify locations in \EditCode{} that may be relevant to editing \OldComment{}.

\noindent\textbf{Features specific to \OldComment{}:}
We include whether a token matches a code token that is inserted, deleted, or replaced in \EditCode{}. These help align parts of \OldComment{} with code edits, assisting the model in determining where edits should be made. In order to exploit common patterns for different types of tokens, we incorporate features that identify whether the token appears more than once in \OldComment{} or is a stop word, and its part-of-speech.

\noindent\textbf{Shared features:}
We include whether the token is a subtoken that was originally part of a larger token and its index if so (e.g., split from \texttt{camelCase}, \texttt{camel} and \texttt{case} are subtokens with indices 0 and 1 respectively). These features aim to encode important relationships between adjacent tokens that are lost once the body of code and comment are transformed into a single, subtokenized sequences. Additionally, because we focus on \Return{} comments, we introduce features intended to guide the model in identifying relevant tokens in \EditCode{} and \OldComment{}. Namely, we include whether a given token matches a token in a \texttt{return} statement that is unique to \OldCode{}, unique to \NewCode{}, or present in both. Similarly, we indicate whether the token matches a token in the subtokenized \texttt{return} type that is unique to \OldCode{}, unique to \NewCode{}, or present in both.

\section{Reranking}
\vspace{-5pt}

Reranking allows the incorporation of additional priors that are difficult to back-propagate, by re-scoring candidate sequences during beam search~\cite{neubig-etal-2015-neural, ko-etal-2019-linguistically, kriz-etal-2019-complexity}.
We incorporate two heuristics to re-score the candidates: 1) generation likelihood and 2) similarity to \OldComment{}. These heuristics are computed after parsing the candidate edit sequences 
(\S\ref{sec:post-processing}).

\noindent\textbf{Generation likelihood.}
Since the \editmodel is trained on edit actions only, it does not globally score the resulting comment in terms of aspects such as fluency and overall suitability for the updated method.
To this end, we make use of a pre-trained comment generation model (\S\ref{sec:generation-model}) that is trained on a substantial amount of data for generating \NewComment{} given only \NewCode{}. We compute the length-normalized probability of this model generating the parsed candidate comment, \PredictedNewComment{}, (i.e., {\small$P(\PredictedNewComment{}|\NewCode{})^{1/N}$} where $N$ is the number of tokens in \PredictedNewComment{}). This model gives preference to comments that are more likely for \NewCode{} and are more consistent with the general style of comments.\footnote{We attempted to integrate this model into the training procedure of the \editmodel through joint training; however, this deteriorated performance.}

\noindent\textbf{Similarity to \OldComment{}.}
\label{sec:sim-old-comment}
So far, our model is mainly trained to produce accurate edits; however, we also follow intuitions that edits should be minimal (as an analogy, the use of Levenshtein distance in spelling correction).
To give preference to predictions that accurately update the comment with minimal modifications, we use similarity to \OldComment{} as a heuristic for reranking. We measure similarity between the parsed candidate prediction and \OldComment{} using METEOR~\cite{BanerjeeEtAL2005}.

\noindent\textbf{Reranking score.}
The reranking score for each candidate is a linear combination of the original beam score, the generation likelihood, and the similarity to \OldComment{} with coefficients 0.5, 0.3, and 0.2 respectively (tuned on validation data).

\section{Data}
\label{sec:data}

\begin{table}[!t]
\begin{center}
\small
\begin{tabular}{|l|l|rrr|}
\cline{3-5}
\nocell{2} & \bf Train & \bf Valid & \bf Test \\
\cline{2-5}
\nocell{1} & Examples & 5,791 & 712 & 736 \\
\nocell{1} & Projects & 526 & 274 & 281 \\
\nocell{1} & Edit Actions & 8,350 & 1,038 & 1,046 \\
\cline{2-5}
\nocell{1} & Sim (\OldCode{}, \NewCode{}) & 0.773 & 0.778 & 0.759 \\
\nocell{1} & Sim (\OldComment{}, \NewComment{}) & 0.623 & 0.645 & 0.635 \\
\hline
\multirow{3}{*}{\bf Code}
& Unique & 7,271 & 2,473 & 2,690 \\
& Mean & 86.4 & 87.4 & 97.4 \\
& Median & 46 & 49 & 50 \\
\hline
\multirow{3}{*}{\bf Comm.}
& Unique & 4,823 & 1,695 & 1,737 \\
& Mean & 10.8 & 11.2 & 11.1 \\
& Median & 8 & 9 & 9 \\

\hline
\end{tabular}
\end{center}
\vspace{-12pt}
\caption{\label{table:partition-stats}\small Number of examples, projects, and edit actions; average similarity between \OldCode{} and \NewCode{} as the ratio of overlap; average similarity between \OldComment{} and \NewComment{} as the ratio of overlap; number of unique code tokens and mean and median number of tokens in a method; and number of unique comment tokens and mean and median number of tokens in a comment.}
\end{table}

We extracted \emph{examples} from popular, open-source Java projects using GitHub's commit history. We extract pairs of the form (method, comment) for the same method across two consecutive commits where there is a simultaneous change to both the code and comment. 
This creates somewhat noisy data for the task of comment update; Appendix~\ref{appendix:filtering} describes filtering techniques to reduce this noise.
We first tokenize \OldCode{} and \NewCode{} using the javalang\footnote{\url{https://pypi.org/project/javalang/}} library. We subtokenize based on camelCase and snake\_case,  as in previous work~\cite{allamanis2016convolutional, Alon2019Code2Seq, Fernandes2019StructuredNeural}. We then form \EditCode{} from the subtokenized forms of \OldCode{} and \NewCode{}.
We tokenize \OldComment{} and \NewComment{} by splitting by space and punctuation. We remove HTML tags and the ``@return''  that precedes all comments, and also subtokenize tokens since code tokens may appear in comments as well. The gold edit action sequence, \EditComment{}, is computed from these processed forms of \OldComment{} and \NewComment{}.

To avoid having examples that closely resemble one another in training and test, 
the projects in the training, test, and validation sets are disjoint, similar to ~\newcite{Movshovitz-AttiasCohen13PredictingProgrammingComments}. 
Table~\ref{table:partition-stats} gives dataset statistics. Of the 7,239 examples in our final dataset, 833 of them were extracted from the diffs used in~~\newcite{panthaplackel2020associating}. Including code and comment tokens that appear at least twice in the training data as well as the predefined edit keywords, the code and comment vocabulary sizes are \CodeVocabSize{} and \CommentVocabSize{} respectively.

\section{Experimental Method}
We evaluate our approach against multiple rule-based baselines and comment generation models. 

\subsection{Baselines}
\noindent\textbf{Copy:} Since much of the content of \OldComment{} is typically retained in the update, we include a baseline that merely copies \OldComment{} as the prediction for \NewComment{}.

\noindent\textbf{Return type substitution:} The return type of a method often appears in its \Return{} comment. If the return type of \OldCode{} appears in \OldComment{} and the return type is updated in the code, we substitute the new return type while copying all other parts of \OldComment{}. Otherwise, \OldComment{} is copied as the prediction.

\noindent\textbf{Return type substitution w/ null handling:} As an addition to the previous method,  we also check whether the token \texttt{null} is added to either a \texttt{return} statement or \texttt{if} statement in the code. If so, we copy \OldComment{} and append the string \textit{or null if null}, otherwise, we simply copy \OldComment{}. This baseline addresses a pattern we observed in the data in which ways to handle \texttt{null} input or cases that could result in \texttt{null} output were added.

\subsection{Generation Model}
\label{sec:generation-model}
One of our main hypotheses is that modeling edit sequences is better suited for this task than generating comments from scratch. However, a counter argument could be that a comment generation model could be trained from substantially more data, since it is much easier to obtain parallel data in the form (method, comment), without the constraints of simultaneous code/comment edits. Hence the power of large-scale training could out-weigh \editmodeling. To this end, we compare with a generation model trained on \GenTrainSet{} method/\Return{} comment pairs collected from GitHub.

We use the same underlying neural architecture as our \editmodel to make sure that the difference in results comes from the amount of training data and from using edit of representations only: 
a two-layer, bi-directional GRU that encodes the sequence of tokens in the method, and an attention-based GRU decoder with a copy mechanism that decodes a sequence of comment tokens. 
We expect the incorporation of more complicated architectures, e.g.,  tree-based~\cite{Alon2019Code2Seq} and graph-based~\cite{Fernandes2019StructuredNeural} encoders which exploit AST structure, can be applied to both an \editmodel and a generation model, which we leave for future work.

Evaluation is based on the \TestSet{} (\NewCode{}, \NewComment{}) pairs in the test set described in \S\ref{sec:data}. 
We ensure that the projects from which training examples are extracted are disjoint from those  in the test set.

\subsection{Reranked Generation Model}
In order to allow the generation model to exploit the old comment, this system uses similarity to \OldComment{} (cf. \S\ref{sec:sim-old-comment}) as a heuristic for reranking the top candidates from the previous model. The reranking score is a linear combination of the original beam score and the METEOR score between the candidate prediction and \OldComment{}, both with coefficient 0.5 (tuned on validation data). 

\subsection{Model Training}
Model parameters are identical across the \editmodel and generation model, tuned on validation data. Encoders have hidden dimension 64, the decoder has hidden dimension 128, and the dimension for code and comment embeddings is 64. The embeddings used in the \editmodel are initialized using the pre-trained embedding vectors from the generation model. We use a dropout rate of 0.6, a batch size of 100, an initial learning rate of 0.001, and Adam optimizer. Models are trained to minimize negative log likelihood, and we terminate training if the validation loss does not decrease for ten consecutive epochs. During inference, we use beam search with beam width=\BeamWidth{}.

\section{Evaluation}

\subsection{Automatic Evaluation}

\begin{table*}[t]
\begin{center}
\small
\begin{tabular}{|l|l|lllll|}
\hline
\bf & \bf Model & \bf xMatch (\%) & \bf METEOR & \bf BLEU-4 & \bf SARI & \bf GLEU \\
\hline 
\multirow{3}{*}{Baselines} & Copy & 0.000 & 34.611 & 46.218 & 19.282 & 35.400 \\
& Return type subt. & 13.723$\ssymbol{4}$ & 43.106$\ssymbol{5}$ & 50.796$\ssymbol{6}$ & 31.723 & 42.507$\ssymbol{1}$ \\
& Return type subst. + null & 13.723$\ssymbol{4}$ & 43.359 & \bf 51.160$\ssymbol{2}$ & 32.109 & 42.627$\ssymbol{1}$ \\
\hline 
\multirow{2}{*}{Models} & Generation & 1.132 & 11.875 & 10.515& 21.164 & 17.350 \\
& Edit & 17.663 & 42.222$\ssymbol{5}$ & 48.217 & {\bf 46.376} & 45.060 \\
\hline 
\multirow{2}{*}{Reranked models} & Generation & 2.083 & 18.170 & 18.891 & 25.641 & 22.685\\
& Edit & {\bf 18.433} & {\bf 44.698} & 50.717$\ssymbol{6}$$\ssymbol{2}$ & 45.486 & {\bf 46.118} \\
\hline
\end{tabular}
\end{center}
\vspace{-12pt}
\caption{\label{table:all-models-nlp}\small Exact match, METEOR, BLEU-4, SARI, and GLEU scores. Scores for which the difference in performance is \textit{not} statistically significant (p \textless{} 0.05) are indicated with matching symbols.}
\end{table*}

\paragraph{Metrics:}
We compute exact match, i.e., the percentage of examples for which the model prediction is identical to the reference comment \NewComment{}. This is often used to evaluate tasks involving source code edits~\cite{ProgramRepairShin, yin19iclr}. We also report two prevailing language generation metrics: METEOR~\cite{BanerjeeEtAL2005}, and average sentence-level BLEU-4~\cite{papineni2002bleu} that is previously used in code-language tasks~\cite{iyer2016acl, LoyolaETAL17GeneratingDescriptionFromSourceCodeChanges}.

Previous work suggests that BLEU-4 fails to accurately capture performance for tasks related  to edits, such as text simplification~\cite{xu-etal-2016-optimizing}, grammatical error correction~\cite{napoles-etal-2015-ground}, and style transfer~\cite{sudhakar-etal-2019-transforming}, since a system that merely copies the input text often achieves a high score.
Therefore, we also include two text-editing metrics to measure how well our system learns to \emph{edit}: SARI~\cite{xu-etal-2016-optimizing}, originally proposed to evaluate text simplification, is essentially the average of N-gram F1 scores corresponding to add, delete, and keep edit operations;\footnote{Although the original formulation only used precision for the delete operation, more recent work computes F1 for this as well~\cite{dong-sentence-simplification, easse-sentence-simplification}.} GLEU~\cite{napoles-etal-2015-ground}, used in grammatical error correction and style transfer, takes into account the source sentence and deviates from BLEU by giving more importance to n-grams that have been correctly changed.

\paragraph{Results:}
We report automatic metrics averaged across three random initializations for all learned models, and use bootstrap tests~\cite{berg-kirkpatrick-etal-2012-empirical} for statistical significance. Table~\ref{table:all-models-nlp} presents the results. While reranking using \OldComment{} appears to help the generation model, it still substantially underperforms all other models, across all metrics. Although this model is trained on considerably more data, it does not have access to \OldComment{} during training and uses fewer inputs and consequently has less context than the \editmodel. Reranking slightly deteriorates the \editmodel's performance with respect to SARI; however, it provides statistically significant improvements on most other metrics.

Although two of the baselines achieve slightly higher BLEU-4 scores than our best model, these differences are not statistically significant, and our model is better at \textit{editing} comments, as shown by the results on exact match, SARI, and GLEU. In particular, our \editmodels beat all other models with wide, statistically significant, margins
on SARI, which explicitly measures performance on edit operations. Furthermore, merely copying \OldComment{}, yields a relatively high BLEU-4 score of 46.218. The \textit{return type substitution} and \textit{return type substitution w/ null handling} baselines produce predictions that are identical to \OldComment{} for 74.73\% and 65.76\% of the test examples, respectively, while it is only 9.33\% for the reranked \editmodel. In other words, the baselines attain high scores on automatic metrics and even beat our model on BLEU-4, without actually performing edits on the majority of examples. This further underlines the shortcomings of some of these metrics and the importance of conducting human evaluation for this task.

\subsection{Human Evaluation}
Automatic metrics often fail to incorporate semantic meaning and sentence structure in evaluation as well as accurately capture performance when there is only one gold-standard reference; indeed, these metrics do not align with human judgment in other generation tasks like grammatical error correction~\cite{napoles-etal-2015-ground} and dialogue generation~\cite{liu-etal-2016-evaluate}.
Since automatic metrics have not yet been explored in the context of the new task we are proposing, we find it necessary to conduct human evaluation and study whether these metrics are consistent with human judgment.

\paragraph{User study design:}
Our study aims to reflect how a comment update system would be used in practice, such as in an Integrated Development Environment (IDE). When developers change code, they would be shown suggestions for updating the existing comment. If they think the comment needs to be updated to reflect the code changes, they could select the one that is most suitable for the new version of the code or edit the existing comment themselves if none of the options are appropriate.

We simulated this setting by asking a user to select the most appropriate  updated comment from a list of suggestions, given \OldComment{} as well as the \textit{diff} between \OldCode{} and \NewCode{} displayed using GitHub's diff interface. The user can select multiple options if they are equally good or a separate \textit{None} option if no update is needed or all suggestions are poor. 

The list of suggestions consists of up to three comments, predicted by the strongest benchmarks and our model
: (1) return type substitution w/ null handling, (2) reranked generation model, and (3) reranked \editmodel, arranged in randomized order. We collapse identical predictions into a single suggestion and reward all associated models if the user selects that comment. Additionally, we remove any prediction that is identical to \OldComment{} to avoid confusion as the user should never select such a suggestion. 
We excluded 
6 examples from the test set for which all three models predicted \OldComment{} for the updated comment.

Nine students (8 graduate/1 undergraduate) and one full-time developer at a large software company, all with 2+ years of Java experience, participated in our study. To measure inter-annotator agreement, we ensured that every example was evaluated by two users. We conducted a total of 500 evaluations, across 250 distinct test examples.

\paragraph{Results:}

\begin{table}[t]
\begin{center}
\small
\begin{tabular}{|cccc|}
\hline
\bf Baseline & \bf Generation & \bf Edit & \bf None \\
\hline
18.4\% & 12.4\% & 30.2\% & 55.0\%\\

\hline
\end{tabular}
\end{center}
\vspace{-12pt}
\caption{\label{table:human-eval-stats}\small Percentage of annotations for which users selected comment suggestions produced by each model. All differences are statistically significant (p \textless{} 0.05).}
\end{table}

Table~\ref{table:human-eval-stats} presents the percentage of annotations (out of 500) for which users selected comment suggestions that were produced by each model. Using Krippendorff's $\alpha$~\cite{alphaAgreement} with MASI distance ~\cite{passonneau-2006-measuring} (which accommodates our multi-label setting), inter-annotator agreement is 0.64, indicating satisfactory agreement.  The reranked \editmodel beats the strongest baseline and reranked generation by wide statistically-significant margins. 
From rationales provided by two annotators, we observe that some options were not selected because they removed relevant information from the existing comment, and not surprisingly, these options often corresponded to the comment generation model.

Users selected none of the suggested comments 55\% of the time, indicating there are many cases for which either the existing comment did not need updating, or comments produced by all models were poor. Based on our inspection of a sample these, we observe that in a large portion of these cases, the comment did not warrant an update. This is consistent with prior work in sentence simplification which shows that, very often, there are sentences that do not need to be simplified~\cite{LiSimplification}.
Despite our efforts to minimize such cases in our dataset through rule-based filtering techniques, we found that many remain. This suggests that it would be beneficial to train a classifier that first determines whether a comment needs to be updated before proposing a revision. Furthermore, the cases for which the existing comment does need to be updated but none of the models produce reasonable predictions illustrate the scope for improvement for our proposed task.

\newsavebox\boxOldGetComplex
\begin{lrbox}{\boxOldGetComplex}
  \begin{lstlisting}[language=java-pretty]
(*@{\sffamily /**{\ttfamily\bf @return} item in {\color{red!60!black}given} position*/}@*)
public Complex getComplex((-W<)final int i(>W-)) {
    return get((-W<)i(>W-));
}
  \end{lstlisting}
\end{lrbox}

\newsavebox\boxNewGetComplex
\begin{lrbox}{\boxNewGetComplex}
  \begin{lstlisting}[language=java-pretty]
(*@{\sffamily /**{\ttfamily\bf @return} item in {\color{green!60!black}first} position*/}@*)
public Complex getComplex() {
    return get();
}
  \end{lstlisting}
\end{lrbox}

\begin{figure}[t]
\centering
\begin{tikzpicture}[
  line width=0.4pt,
  node distance=0ex and 0em,
  every node/.style={scale=1},
  gridBox/.style={rectangle, opacity=0, draw=red},
  box/.style={rectangle, draw=black, inner sep=2pt, font=\small},
  rounded box/.style={rectangle, rounded corners, draw=black, inner sep=2pt, font=\small},
  anno/.style={font=\footnotesize},
]

  \DefMacro{wCodeBox}{20em}
  \DefMacro{hCodeBox}{7.5ex}
  
  \node (g-Old) at (0,0) [box, minimum width=\UseMacro{wCodeBox}, minimum height=\UseMacro{hCodeBox}] {};
  \node (b-Old) [right = 0 of g-Old.west] [] {\usebox\boxOldGetComplex};
  \node (b-OldText) [below left = 1pt and 1pt of g-Old.north east] [box, draw=red!50, fill=red!20, scale=0.8] {\bf Previous Version\phantom{p\hskip -.6em}};
  \node (g-New) [below = 0 of g-Old.south] [box, minimum width=\UseMacro{wCodeBox}, minimum height=\UseMacro{hCodeBox}] {};
  \node (b-New) [right = 0 of g-New.west] [] {\usebox\boxNewGetComplex};
  \node (b-NewText) [below left = 1pt and 1pt of g-New.north east] [box, draw=green!50, fill=green!20, scale=0.8] {\bf Updated Version};
  
\end{tikzpicture}
\vspace{-18pt}
\caption{\small Changes in the \texttt{getComplex} method  and its corresponding  \Return{} comment between two subsequent commits of the eclipse-january project, available on GitHub.}
\label{fig:eclipse-january}
\vspace{-15pt}
\end{figure}

\section{Error Analysis}
We find that our model performs poorly in cases requiring external knowledge and more context than that provided by the given method. For instance, correctly updating the comment shown in Figure~\ref{fig:eclipse-january} requires knowing that \texttt{get} returns the item in the first position if no argument is provided. Our model does not have access to this information, and it fails to generate a reasonable update: ``\Return{} complex in given position." On the other hand, the reranked generation model produces ``\Return{} the complex value" which is arguably reasonable for the given context. This suggests that incorporating more code context could be beneficial for both models. Furthermore, we find that our model tends to make more mistakes when it must reason about a large amount of code change between \OldCode{} and \NewCode{}, and we found that in many such cases, the output of the reranked generation model was better. This suggests that when there are substantial code changes, \NewCode{} effectively becomes a \textit{new} method, and generating a comment from scratch may be more appropriate. Ensembling generation with our system through a regression model that predicts the extent of editing that is needed may lead to a more generalizable approach that can accommodate such cases. Sample outputs are given in Appendix~\ref{appendix:sample_output}.

\begin{table*}[htb]
\begin{center}
\small
\begin{tabular}{|l|l|lllll|}
\hline
\bf Inputs & \bf Output & \bf xM (\%) & \bf METEOR  & \bf BLEU-4 & \bf SARI & \bf GLEU \\
\hline
\multirow{2}{*}{\OldComment{}, \NewCode{}} & \NewComment{} & 5.707$\ssymbol{3}$$\ssymbol{5}$ & 29.259$\ssymbol{2}$ & 33.534$\ssymbol{4}$ & 28.024 & 30.000$\ssymbol{1}$\\
& \EditComment{} & 4.755$\ssymbol{3}$$\ssymbol{1}$ & 33.796 & 43.315 & 35.516 & 37.970$\ssymbol{6}$ \\
\hline
\multirow{2}{*}{\OldComment{}, \OldCode{}, \NewCode{}} & \NewComment{} & 3.714$\ssymbol{1}$ & 18.729 & 20.060 & 23.914 & 21.956\\
& \EditComment{} & 5.163$\ssymbol{3}$$\ssymbol{5}$ & 34.895 & 44.006$\ssymbol{1}$ & 33.479 & 37.618$\ssymbol{6}$ \\
\hline
\multirow{2}{*}{\OldComment{}, \EditCode{}} & \NewComment{} & 6.114$\ssymbol{5}$ & 29.968$\ssymbol{2}$ & 34.164$\ssymbol{4}$ & 28.980 & 30.491$\ssymbol{1}$ \\
& \EditComment{} & {\bf 8.922} & {\bf 36.229} & {\bf 44.283}$\ssymbol{1}$ & {\bf 40.538} & {\bf 39.879}\\
\hline
\end{tabular}
\end{center}
\vspace{-12pt}
\caption{\label{table:edit-reps} \small Exact match, METEOR, BLEU-4, SARI, and GLEU for various combinations of code input and target comment output configurations. Features and reranking are disabled for all models. Scores for which the difference in performance is \textit{not} statistically significant (p \textless{} 0.05) are indicated with matching symbols.}
\end{table*}

\begin{table*}[h]
\begin{center}
\small
\begin{tabular}{|l|l|lllll|}
\hline
& \bf Model & \bf xM (\%) & \bf METEOR & \bf BLEU-4 & \bf SARI & \bf GLEU \\
\hline 
\multirow{2}{*}{Models} & Edit & 17.663 & 42.222 & 48.217 & {\bf 46.376} & 45.060 \\
 & - feats. & 8.922$\ssymbol{2}$ & 36.229 & 44.283 & 40.538 & 39.879$\ssymbol{1}$ \\
\hline 
\multirow{2}{*}{Reranked models} & Edit & {\bf 18.433} & {\bf 44.698} & {\bf 50.717} & 45.486 & {\bf 46.118}\\
 & - feats. & 8.877$\ssymbol{2}$ & 38.446 & 46.665 & 36.924 & 40.317$\ssymbol{1}$ \\
\hline
\end{tabular}
\end{center}
\vspace{-12pt}
\caption{\label{table:feature-ablations}\small Exact match, METEOR, BLEU-4, SARI, and GLEU scores of ablated models. Scores for which the difference in performance is \textit{not} statistically significant (p \textless{} 0.05) are indicated with matching symbols.}
\end{table*}

\section{Ablations}
\label{appendix:ablations}
We empirically study the effect of training the network to encode explicit code edits and decode explicit comment edits. As discussed in Section~\ref{sec:edit-model}, the \editmodel consists of two encoders, one that encodes \OldComment{} and another that encodes the code representation, \EditCode{}. We conduct experiments in which the code representation instead consists of either (1) \NewCode{} or (2) both \OldCode{} and \NewCode{} (encoded separately and hidden states concatenated). Additionally, rather than having the decoder generate comment edits in the form \EditComment{}, we introduce experiments in which it directly generates \NewComment{}, with no intermediate edit sequence. For this, we use only the underlying architecture of the \editmodel (without features or reranking). The performance for various combinations of input code and target comment representations are shown in Table~\ref{table:edit-reps}.

By comparing performance across combinations consisting of the same input code representation and varying target comment representations, the importance of training the decoder to generate a sequence of edit actions rather than the full updated comment is very evident. Furthermore, comparing across varying code representations under the \EditComment{} target comment representation, it is clear that explicitly encoding the code changes, as \EditCode{}, leads to significant improvements across most metrics.

We further ablate the features introduced in \S\ref{sec:features}. As shown in Table~\ref{table:feature-ablations}, these features improve performance by wide margins, across all metrics.

\section{Related Work}

\noindent \textbf{Learning from source code changes:}
~\newcite{apiUpdate} use rule-based techniques to automatically detect and revise outdated API names in code documentation; however, their approach cannot be extended to full natural language comments that are the focus of this work. ~\newcite{ZhaiCPC} propose a technique for updating incomplete and buggy comments by propagating comments from different code elements (e.g., variables, methods, classes) based on program analysis and several heuristics. Rather than simply copying a related comment, we aim to revise an outdated comment by reasoning about code changes.~\newcite{yin19iclr} present an approach for learning structural and semantic properties of source code edits so that they can be generalized to new code inputs. Similar to their work, we learn vector representations from source code changes; however, unlike their setting, we apply these representations to natural language. Prior work in automatic commit message generation aims to learn from code changes in order to generate a natural language summary of these changes~\cite{LoyolaETAL17GeneratingDescriptionFromSourceCodeChanges, JiangCommit, XuCommit}. Instead of generating natural language content from scratch as done in their work, we focus on applying edits to existing natural language text. We also show that generating a comment from scratch does not perform as well as our proposed \editmodel for the comment update setting.

\noindent \textbf{Editing natural language text:}
Approaches for editing natural language text have been studied extensively through tasks such as sentence simplification~\cite{dong-sentence-simplification}, style transfer~\cite{li-style-transfer}, grammatical error correction~\cite{awasthi-etal-2019-parallel}, and language modeling~\cite{Guu2018GeneratingSB}. 
The focus of this prior work is to revise sentences to conform to stylistic and grammatical conventions, and it does not generally consider broader contextual constraints. On the contrary, our goal is not to make cosmetic revisions to a given span of text, but rather amend its semantic meaning to be in sync with the content of a separate body of information on which it is dependent. More recently, ~\newcite{Shah2020AutomaticFS} proposed an approach for rewriting an outdated sentence based on a sentence stating a new factual claim, which is more closely aligned with our task. However, in our case, the separate body of information is not natural language and is generally much longer than a single sentence.

\section{Conclusion}
We have addressed the novel task of automatically updating an existing programming comment based on changes to the related code. We designed a new approach for this task which aims to correlate cross-modal edits in order to generate a sequence of edit actions specifying how the comment should be updated. We find that our model outperforms multiple rule-based baselines and comment generation models, with respect to several automatic metrics and human evaluation.

\section*{Acknowledgements}
We thank reviewers for their feedback on this work and the
participants of our user study for their time. This work was partially
supported by a Google Faculty Research Award and the US National
Science Foundation under Grant Nos.~CCF-1652517 and IIS-1850153.

\bibliography{acl2020}
\bibliographystyle{acl_natbib}

\clearpage
\newpage
\appendix

\section{Modified Comment Edit Lexicon}
\label{appendix:comment-edit-lexicon}

\begin{table}[!t]
\begin{center}
\small
\begin{tabular}{|l|rrr|}
\hline
 & \bf Train & \bf Valid & \bf Test \\
\hline
Total actions & 8,350 & 1,038 & 1,046 \\
Avg. \# actions per example & 1.44 & 1.46 & 1.42 \\
\hline
\Replace{} & 51.9\% & 49.7\% & 50.1\% \\
\ReplaceKeepBefore{} & 2.9\% & 2.6\% & 3.5\% \\
\ReplaceKeepAfter{} & 0.7\% & 0.3\% & 0.4\% \\
\hline
\InsertKeepBefore{} & 21.5\% & 24.1\% & 23.2\% \\
\InsertKeepAfter{} & 4.2\% & 4.0\% & 3.3\% \\
\hline
\Delete{} & 17.4\% & 18.0\% & 17.8\% \\
\DeleteKeepBefore{} & 1.3\% & 0.7\% & 1.1\% \\
\DeleteKeepAfter{} & 0.2\% & 0.5\% & 0.6\%  \\
\hline

\hline
\end{tabular}
\end{center}
\vspace{-12pt}
\caption{\label{table:edit-action-stats}\small Total number of edit actions; average number of edit actions per example; percentage of total actions that is accounted by each edit action type.}
\end{table}

We first transform insertions and ambiguous deletions into a structure that resembles \Replace{}, characterized by \InsertOld{}/\InsertNew{} and \DeleteOld{}/\DeleteNew{} spans respectively. Next, we require the span of tokens attached to \ReplaceOld{}, \InsertOld{}, and \DeleteOld{} to be unique across \OldComment{} so that we can uniquely identify the edit location. We enforce this by iteratively searching through unchanged tokens before and after the span, incorporating additional tokens into the span, until the span becomes unique. These added tokens are then included in both components of the action. For instance, if the last \texttt{A} is to be replaced with \texttt{C} in \texttt{ABA}, the \ReplaceOld{} span would be \texttt{BA} and the \ReplaceNew{} span would be \texttt{BC}. We also augment the edit types to differentiate between the various scenarios that may arise from this search procedure.

\Replace{} actions for which this procedure is performed deviate from the typical nature of \Replace{} in which there is no overlap between the spans attached to \ReplaceOld{} and \ReplaceNew{}. This is because the tokens that are added to make the \ReplaceOld{} span unique will appear in both spans. These tokens, which are effectively kept between \OldComment{} and \NewComment{}, could appear before or after the edit location. We differentiate between these scenarios by augmenting the edit lexicon with new edit types. In addition to \Replace{}, we have \ReplaceKeepBefore{} and \ReplaceKeepAfter{} to signify that the action entails retaining some content before or after, respectively. We include the same for the other types as well with \InsertKeepBefore{}, \InsertKeepAfter{}, \DeleteKeepBefore{}, \DeleteKeepAfter{}. Table~\ref{table:edit-action-stats} shows statistics on how often each of these edit actions are used. We present more details about the actions in the sections that follow.

\subsection{Replacements}
\paragraph{\Replace{}} This action is defined as shown below:
\noindent
\begin{equation*}
  \begin{array}{l}
    \texttt{\small <ReplaceOld>[span of old tokens]}\\
    \noalign{\vskip-6pt}
    \texttt{\small <ReplaceNew>[span of new tokens]}\\
    \noalign{\vskip-6pt}
    \texttt{\small <ReplaceEnd>}\\
    \noalign{\vskip-6pt}
  \end{array}
\end{equation*}
It prescribes that the tokens attached to \ReplaceOld{} are deleted and the tokens attached to \ReplaceNew{} are inserted in their place. There is almost never overlap between the span of tokens attached to \ReplaceOld{} and \ReplaceNew{}. Example: if \texttt{B} is to be replaced with \texttt{C} in \OldComment{}=\texttt{AB} to produce \NewComment{}=\texttt{AC}, the corresponding \EditComment{} is:
\noindent
\begin{equation*}
  \begin{array}{l}
    \texttt{\small <ReplaceOld>B}\\
    \noalign{\vskip-6pt}
    \texttt{\small <ReplaceNew>C}\\
    \noalign{\vskip-6pt}
    \texttt{\small <ReplaceEnd>}\\
    \noalign{\vskip-6pt}
  \end{array}
\end{equation*}
Note that the span attached to \ReplaceOld{} must be unique across \OldComment{} for this edit type to  be used.

\paragraph{\ReplaceKeepBefore{}} This action is defined as shown below:
\noindent
\begin{equation*}
  \begin{array}{l}
    \texttt{\small <ReplaceOldKeepBefore>[span of old tokens]}\\
    \noalign{\vskip-6pt}
    \texttt{\small <ReplaceNewKeepBefore>[span of new tokens]}\\
    \noalign{\vskip-6pt}
    \texttt{\small <ReplaceEnd>}\\
    \noalign{\vskip-6pt}
  \end{array}
\end{equation*}
\Replace{} is transformed into this structure if the span attached to \ReplaceOld{} is not unique. For example, suppose the first \texttt{B} is to be replaced with \texttt{D} in \OldComment{}=\texttt{ABCB} to produce \NewComment{}=\texttt{ADCB}. If \EditComment{} consists of a \ReplaceOld{} span carrying just \texttt{B}, it is not obvious whether the first or last \texttt{B} should be replaced. To address this, we introduce a new edit type, \ReplaceKeepBefore{}, which forms a unique span by searching before the edit location. It prescribes that the tokens attached to \ReplaceOldKeepBefore{} are deleted and the tokens attached to \ReplaceNewKeepBefore{} are inserted in their place. Unlike \Replace{}, there will be some overlap at the beginning of the spans attached to \ReplaceOldKeepBefore{} and \ReplaceNewKeepBefore{}. Therefore, to represent editing \OldComment{}=\texttt{ABCB} to produce \NewComment{}=\texttt{ADCB}, \EditComment{} is:
\noindent
\begin{equation*}
  \begin{array}{l}
    \texttt{\small <ReplaceOldKeepBefore> AB}\\
    \noalign{\vskip-6pt}
    \texttt{\small <ReplaceNewKeepBefore> AD}\\
    \noalign{\vskip-6pt}
    \texttt{\small <ReplaceEnd>}\\
    \noalign{\vskip-6pt}
  \end{array}
\end{equation*}
The span attached to \ReplaceOldKeepBefore{} is unique, making it clear that the first \texttt{B} is to be replaced with \texttt{D}. It also indicates that we are effectively keeping \texttt{A}, which appears before the edit location.

\paragraph{\ReplaceKeepAfter{}} This action is defined as shown below:
\noindent
\begin{equation*}
  \begin{array}{l}
    \texttt{\small <ReplaceOldKeepAfter>[span of old tokens]}\\
    \noalign{\vskip-6pt}
    \texttt{\small <ReplaceNewKeepAfter>[span of new tokens]}\\
    \noalign{\vskip-6pt}
    \texttt{\small <ReplaceEnd>}\\
    \noalign{\vskip-6pt}
  \end{array}
\end{equation*}
\Replace{} is transformed into this structure if the span attached to \ReplaceOld{} is not unique and \ReplaceKeepBefore{} cannot be used because we are unable to find a unique sequence of unchanged tokens before the edit location. For example, suppose the first \texttt{B} is to be replaced with \texttt{D} in \OldComment{}=\texttt{ABCAB} to produce \NewComment{}=\texttt{ADCAB}. Searching before the edit location, we find only \texttt{AB}, which is not unique across \OldComment{}, and so it would still not be clear which \texttt{B} is to be edited. To address this, we introduce a new edit type, \ReplaceKeepAfter{}, which forms a unique span by searching \textit{after} the edit location. It prescribes that the tokens attached to \ReplaceOldKeepAfter{} are deleted and the tokens attached to \ReplaceNewKeepAfter{} are inserted in their place. Unlike \Replace{} and \ReplaceKeepBefore{}, there will be some overlap at the end of the spans attached to \ReplaceOldKeepAfter{} and \ReplaceNewKeepAfter{}. Therefore, to represent editing \OldComment{}=\texttt{ABCAB} to produce \NewComment{}=\texttt{ADCAB}, \EditComment{} is:
\noindent
\begin{equation*}
  \begin{array}{l}
    \texttt{\small <ReplaceOldKeepAfter> BC}\\
    \noalign{\vskip-6pt}
    \texttt{\small <ReplaceNewKeepAfter> DC}\\
    \noalign{\vskip-6pt}
    \texttt{\small <ReplaceEnd>}\\
    \noalign{\vskip-6pt}
  \end{array}
\end{equation*}
The span attached to \ReplaceOldKeepAfter{} is unique, making it clear that the first \texttt{B} is to be replaced with \texttt{D}. It also indicates that we are effectively keeping \texttt{C}, which appears after the edit location.

\subsection{Insertions}
We disregard basic \Insert{} actions since it is always ambiguous where an insertion should occur without an anchor point. Following what is done for ambiguous \Replace{} actions, we introduce \InsertKeepBefore{} and \InsertKeepAfter{}.

\paragraph{\InsertKeepBefore{}} This action is defined as shown below:
\noindent
\begin{equation*}
  \begin{array}{l}
    \texttt{\small <InsertOldKeepBefore>[span of old tokens]}\\
    \noalign{\vskip-6pt}
    \texttt{\small <InsertNewKeepBefore>[span of new tokens]}\\
    \noalign{\vskip-6pt}
    \texttt{\small <InsertEnd>}\\
    \noalign{\vskip-6pt}
  \end{array}
\end{equation*}
In this representation, the span of tokens attached to \InsertOldKeepBefore{} must be unique and serve as the anchor point for where the new tokens should be inserted. We do this by searching before the edit location. The structure is identical to that of \ReplaceKeepBefore{} in that the tokens attached to \InsertOldKeepBefore{} are replaced with the tokens in \InsertNewKeepBefore{} and that there is some overlap at the beginning of the two spans. As an example, suppose \texttt{C} is to be inserted at the end of \OldComment{}=\texttt{AB} to form \NewComment{}=\texttt{ABC}. Then the corresponding \EditComment{} is as follows:
\noindent
\begin{equation*}
  \begin{array}{l}
    \texttt{\small <InsertKeepBefore> B}\\
    \noalign{\vskip-6pt}
    \texttt{\small <InsertNewKeepBefore> BC}\\
    \noalign{\vskip-6pt}
    \texttt{\small <InserteEnd>}\\
    \noalign{\vskip-6pt}
  \end{array}
\end{equation*}
This states that we are effectively inserting \texttt{C} and keeping \texttt{B}, which appears before the edit location.

\paragraph{\InsertKeepAfter{}} This action is defined as shown below:
\noindent
\begin{equation*}
  \begin{array}{l}
    \texttt{\small <InsertOldKeepAfter>[span of old tokens]}\\
    \noalign{\vskip-6pt}
    \texttt{\small <InsertNewKeepAfter>[span of new tokens]}\\
    \noalign{\vskip-6pt}
    \texttt{\small <InsertEnd>}\\
    \noalign{\vskip-6pt}
  \end{array}
\end{equation*}
We rely on this when we are unable to use \InsertKeepBefore{} because we cannot find a unique span of tokens to identify the anchor point, by searching before the edit location. For instance,  suppose \texttt{C} is to be inserted at the beginning of \OldComment{}=\texttt{AB} to form \NewComment{}=\texttt{CAB}. There are no tokens that appear before the insert point, so we instead choose to search \textit{after}. The structure is identical to that of \ReplaceKeepAfter{} in that the tokens attached to \InsertOldKeepAfter{} are replaced with the tokens in \InsertNewKeepAfter{} and that there is some overlap at the end of the two spans. The corresponding \EditComment{} from our example is as follows:
\noindent
\begin{equation*}
  \begin{array}{l}
    \texttt{\small <InsertKeepAfter> A}\\
    \noalign{\vskip-6pt}
    \texttt{\small <InsertNewKeepAfter> CA}\\
    \noalign{\vskip-6pt}
    \texttt{\small <InserteEnd>}\\
    \noalign{\vskip-6pt}
  \end{array}
\end{equation*}
This states that we are effectively inserting \texttt{C} and keeping \texttt{A}, which appears after the edit location.

\subsection{Deletions}
\paragraph{\Delete{}} This action is defined as shown below:
\noindent
\begin{equation*}
  \begin{array}{l}
    \texttt{\small <Delete>[span of old tokens]<DeleteEnd>}\\
    \noalign{\vskip-6pt}
  \end{array}
\end{equation*}
It prescribes that the tokens that appear in the \Delete{} span are removed from \OldComment{}. Example: if \texttt{B} is to be deleted from \OldComment{}=\texttt{AB} to produce \NewComment{}=\texttt{A}, the corresponding \EditComment{} is:
\noindent
\begin{equation*}
  \begin{array}{l}
    \texttt{\small <Delete>B<DeleteEnd>}\\
    \noalign{\vskip-6pt}
  \end{array}
\end{equation*}
Note that the \Delete{} span must be unique across \OldComment{} for this edit type to be used.

\paragraph{\DeleteKeepBefore{}} This action is defined as shown below:
\noindent
\begin{equation*}
  \begin{array}{l}
    \texttt{\small <DeleteOldKeepBefore>[span of old tokens]}\\
    \noalign{\vskip-6pt}
    \texttt{\small <DeleteNewKeepBefore>[span of new tokens]}\\
    \noalign{\vskip-6pt}
    \texttt{\small <DeleteEnd>}\\
    \noalign{\vskip-6pt}
  \end{array}
\end{equation*}
\Delete{} is transformed into this structure if the \Delete{} span is not unique. For example, suppose the first \texttt{B} is to be deleted from \OldComment{}=\texttt{ABCB} to produce \NewComment{}=\texttt{ACB}. From just \EditComment{}=$\texttt{\small<Delete>B<DeleteEnd>}$, it is unclear which \texttt{B} is to be deleted. To address this, we introduce a new edit type, \DeleteKeepBefore{}, which forms a unique span by searching before the edit location. The structure is identical to that of \ReplaceKeepBefore{} in that the tokens attached to \DeleteOldKeepBefore{} are replaced with the tokens in \DeleteNewKeepBefore{} and that there is some overlap at the beginning of the two spans. For the example under consideration, the corresponding \EditComment{} is given below:
\noindent
\begin{equation*}
  \begin{array}{l}
    \texttt{\small <DeleteOldKeepBefore> AB}\\
    \noalign{\vskip-6pt}
    \texttt{\small <DeleteNewKeepBefore> A}\\
    \noalign{\vskip-6pt}
    \texttt{\small <DeleteEnd>}\\
    \noalign{\vskip-6pt}
  \end{array}
\end{equation*}
The span attached to \DeleteOldKeepBefore{} is unique, making it clear that the first \texttt{B} is to be deleted. It also indicates that we are effectively keeping \texttt{A}, which appears before the edit location.

\paragraph{\DeleteKeepAfter{}} This action is defined as shown below:
\noindent
\begin{equation*}
  \begin{array}{l}
    \texttt{\small <DeleteOldKeepAfter>[span of old tokens]}\\
    \noalign{\vskip-6pt}
    \texttt{\small <DeleteNewKeepAfter>[span of new tokens]}\\
    \noalign{\vskip-6pt}
    \texttt{\small <DeleteEnd>}\\
    \noalign{\vskip-6pt}
  \end{array}
\end{equation*}
\Delete{} is transformed into this structure if the \Delete{} span is not unique and \DeleteKeepBefore{} cannot be used because we are unable to find a unique sequence of unchanged tokens before the edit location. For example, suppose the first \texttt{B} is to be deleted from \OldComment{}=\texttt{ABCAB} to produce \NewComment{}=\texttt{ACAB}. Searching before the edit location, we find only \texttt{AB}, which is not unique across \OldComment{}, and so it would still not be clear which \texttt{B} is to be deleted. To address this, we introduce a new edit type, \DeleteKeepAfter{}, which forms a unique span by searching \textit{after} the edit location. The structure is identical to that of \ReplaceKeepAfter{} in that the tokens attached to \DeleteOldKeepAfter{} are replaced with the tokens in \DeleteNewKeepAfter{} and that there is some overlap at the end of the two spans. For the example under consideration, \EditComment{} is given below:
\noindent
\begin{equation*}
  \begin{array}{l}
    \texttt{\small <DeleteOldKeepAfter> BC}\\
    \noalign{\vskip-6pt}
    \texttt{\small <DeleteNewKeepAfter> C}\\
    \noalign{\vskip-6pt}
    \texttt{\small <DeleteEnd>}\\
    \noalign{\vskip-6pt}
  \end{array}
\end{equation*}
The span attached to \DeleteOldKeepAfter{} is unique, making it clear that the first \texttt{B} is to be deleted. It also indicates that we are effectively keeping \texttt{C}, which appears after the edit location.

\section{Data Filtering}
\label{appendix:filtering}
As done in~\newcite{panthaplackel2020associating}, we apply heuristics to reduce the number of cases in which the code and comment changes are unrelated.
First, because we focus on \Return{} comments that pertain to the return values of a given method, we discard any example in which the code change does not entail either a change to the return type or at least one return statement.
Then, to identify the correct mapping of two versions of a method
among other changes in a commit, we focus on the code changes that
preserve the method names.  It may happen sometimes that developers
change the method name as well as code and comment in one commit, but
we leave it as future work to improve this filtering heuristic.
Next, we attempt to remove examples in which the comment change appears to be purely stylistic (e.g. spelling corrections, re-formatting, and rephrasing).
Furthermore, prior work~\cite{allamanis2019duplication} has shown that duplication can adversely affect evaluation of machine learning models for code and language tasks. For this reason, we remove duplicates from our corpus.

Despite having mined commit histories for thousands of projects, upon filtering, we are left with a total of \CorpusSize{} examples belonging to \NumProjects{} different projects. This demonstrates the challenge of collecting large datasets with relatively low levels of noise in this domain. Although online code resources like GitHub and StackOverflow host large quantities of data that can be exploited for transduction tasks between source code and natural language, prior work has shown that much of this data is unusable without cleaning~\cite{yin2018mining}.

Some have used rule-based techniques to do data cleaning~\cite{allamanis2016convolutional, Hu2018DeepCC, Fernandes2019StructuredNeural}, and others train classifiers on hand-labeled examples that can be applied to a much larger pool of examples in order to differentiate between clean and noisy examples~\cite{iyer2016acl, yao2018staqc, yin2018mining}. Most of these approaches focus on code summarization or comment generation which only require single code-NL pairs for training and evaluation as the task entails generating a natural language summary of a given code snippet. On the contrary, our proposed task requires two code-NL pairs that are assumed to hold specific parallel relationships with one another. Namely, the relationship between \NewComment{} and \NewCode{} is expected to be similar to that of \OldComment{} and \OldCode{}. The relationship between \NewComment{} and \OldComment{} is expected to correlate with the relationship between \NewCode{} and \OldCode{}. Not only does having four moving parts in one example magnify noise, but the need to hold these relationships makes data cleaning particularly difficult. We leave building classifiers for aiding this process as future work.

\section{Sample Output}
\label{appendix:sample_output}

\begin{table*}[t]
	\begin{center}
          \includegraphics{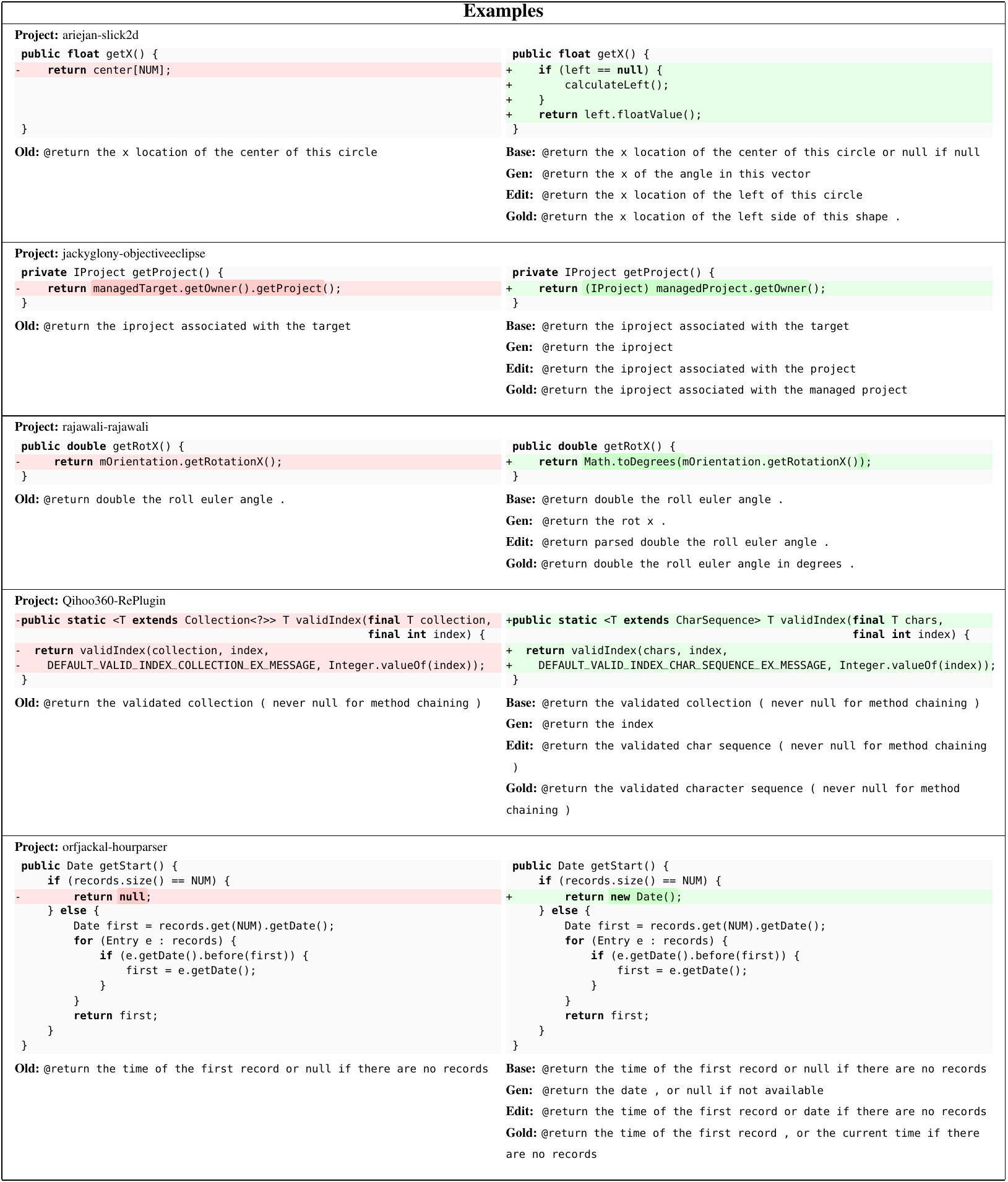}
	\end{center}
	\vspace{-12pt}
	\caption{\small Examples from open-source software projects. For each example, we show the diff between the two versions of the method (left: old version, right: new version, diff lines are highlighted), the existing \Return{} comment prior to being updated (left), and predictions made by the \textit{return type substitution w/ null handling} baseline, reranked generation model, and reranked \editmodel, and the gold updated comment (right, from top to bottom).}
	\label{table:sample_output}
	\vspace{90pt}  %
\end{table*}

In Table~\ref{table:sample_output}, we show predictions for various examples in the test set.

\end{document}